\documentclass[11pt,a4paper,reqno]{amsart}
\usepackage{tikz}
\usepackage{pgfplots}
\usetikzlibrary{calc}
\usepackage{mathtools}
\usepackage{hyperref}

\usepackage[headings]{fullpage}
\usepackage{color,multirow}
\usepackage{url}

\newcommand{\dinit}{\mathsf{init}}

\newcommand{\dsucc}{\mathsf{succ}}

\newcommand{\dresp}{\mathsf{resp}}

\newcommand{\dweakresp}{\mathsf{weakresp}}

\newcommand{\dprec}{\mathsf{prec}}

\newcommand{\dmustexist}{\mathsf{mustexist}}
\newcommand{\dorresp}{\mathsf{orresp}}
\newcommand{\Const}{\mathsf{Const}}

\newcommand{\GETOUT}[1]{}
\definecolor{mygreen}{RGB}{80,160,80}
\definecolor{myonegrey}{RGB}{30,30,30}
\definecolor{mytwogrey}{RGB}{70,70,70}
\definecolor{mythreegrey}{RGB}{110,110,110}
\definecolor{myfourgrey}{RGB}{150,150,150}
\newcommand\mps[1]{}

\newcommand{\valid}{\mathsf{valid}}

\newcommand{\tbox}[1]{\begin{minipage}[t]{1.5cm} \begin{center} \tiny\bf #1\end{center}\end{minipage}}

\newcommand{\myarr}[3]{\draw [haup] (#1) to node[fill=white] {#3} (#2);}

\newcommand*{\doi}[1]{\href{http://dx.doi.org/#1}{doi:#1}}
\def\FDAP{FDAP}
\def\FDAPMone{\FDAP M1}
\def\FDAPMtwo{\FDAP M2}
\def\FDAPMthree{\FDAP M3}

\tikzset{empty/.style={fill=none, minimum height =0mm}}

\newcommand{\femadiagone}{
\begin{figure*}[!h]
\fbox{
\begin{tikzpicture}[outline/.style={draw,thick,fill=black!3, minimum height=0.75cm,minimum width=1.5cm},
                    outline/.default=black,
                    haup/.style={thick,->},
                    hauq/.style={thick,}
                    ]
\def\myscl{2}
\node (N0) at (0,12) {};
\node (A0) at (1*\myscl,9) {};
\node [outline] (N1) at (1*\myscl,12) {\tbox{1. Disaster strikes}};
\node [outline] (N2) at (0,10) {\tbox{2. Disaster identified}};
\node [outline] (N3) at (0,8) {\tbox{3. Damage assessment}};
\node [outline] (N4) at (0,6) {\tbox{4. Gov't review damage}};
\node [outline] (N5) at (0,4) {\tbox{5. Gov'r assess response resources}};
\node [outline] (N7) at (2*\myscl,6) {\tbox{7. Region submits MDD+FA request}};
\node [outline] (N8) at (6*\myscl,6) {\tbox{8. President reviews request}};
\node [outline] (N9) at (6*\myscl,9) {\tbox{9. President declares disaster}};
\node [outline] (N10) at (6*\myscl,12) {\tbox{10. FEMA support delivered}};
\node [outline] (N6) at (2*\myscl,2) {\tbox{6. State only disaster response}};
\draw [hauq] (N1) -- (0,12) -- (N2);
\myarr{N0}{N2}{\tiny$\dprec$}
\myarr{N2}{N3}{\tiny$\dsucc$}
\myarr{N3}{N4}{\tiny$\dsucc$}
\myarr{N4}{N5}{\tiny$\dprec$}
\myarr{N7}{N8}{\tiny$\dsucc$}
\myarr{N8}{N9}{\tiny$\dweakresp$}
\myarr{N9}{N10}{\tiny$\dsucc$}
\draw [hauq] (N1) -- (1*\myscl,9) -- (N9);
\myarr{A0}{N9}{\tiny$\dprec$}
\myarr{N5}{N7}{\tiny$\dprec$}
\myarr{N5}{N6}{\tiny$\dprec$}
\node [draw,circle,minimum width=.3cm,fill=white] (Cen) at (1.3*\myscl,4) {\tiny{orresp}};
\draw [haup] (N5) -- (Cen);
\draw [haup] (Cen) -- (N7);
\draw [haup] (Cen) -- (N6);
\node at (1*\myscl,12.75) {\tiny$\dinit$};
\end{tikzpicture}
}
\caption{The declarative process $D_{\FDAP}$ modelling the FDAP.}
 \label{fema-as-is}
\end{figure*}
}

\newcommand{\femadiagtwo}{
\begin{figure*}[!h]
\fbox{
\begin{tikzpicture}[
					thick,scale=0.6, every node/.style={transform shape},
					outline/.style={draw,thick,fill=black!3, minimum height=0.75cm,minimum width=1.5cm},
                    outline/.default=black,
                    haup/.style={thick,->},
                    hauq/.style={thick,}
                    ]
\def\myscl{2}
\node (N0) at (0,12) {};
\node (A0) at (1*\myscl,9) {};
\node [outline] (N1) at (1*\myscl,12) {\tbox{1. Disaster strikes}};
\node [outline] (N2) at (0,10) {\tbox{2. Disaster identified}};
\node [outline] (N3) at (0,8) {\tbox{3. Damage assessment}};
\node [outline] (N4) at (0,6) {\tbox{4. Gov't review damage}};
\node [outline] (N5) at (0,4) {\tbox{5. Gov'r assess response resources}};
\node [outline] (N7) at (2*\myscl,6) {\tbox{7. Region submits MDD+FA request}};
\node [outline] (N8) at (6*\myscl,6) {\tbox{8. President reviews request}};
\node [outline] (N9) at (6*\myscl,9) {\tbox{9. President declares disaster}};
\node [outline] (N10) at (6*\myscl,12) {\tbox{10. FEMA support delivered}};
\node [outline] (N6) at (2*\myscl,2) {\tbox{6. State only disaster response}};
\draw [hauq] (N1) -- (0,12) -- (N2);
\myarr{N0}{N2}{\tiny$\dprec$}
\myarr{N2}{N3}{\tiny$\dsucc$}
\myarr{N3}{N4}{\tiny$\dsucc$}
\myarr{N4}{N5}{\tiny$\dprec$}
\myarr{N7}{N8}{\tiny$\dsucc$}
\myarr{N8}{N9}{\tiny$\dweakresp,\dprec$}
\myarr{N9}{N10}{\tiny$\dsucc$}
\myarr{N5}{N7}{\tiny$\dprec$}
\myarr{N5}{N6}{\tiny$\dprec$}
\node [draw,circle,minimum width=.3cm,fill=white] (Cen) at (1.3*\myscl,4) {\tiny{orresp}};
\draw [haup] (N5) -- (Cen);
\draw [haup] (Cen) -- (N7);
\draw [haup] (Cen) -- (N6);
\node at (1*\myscl,12.75) {\tiny$\dinit$};
\end{tikzpicture}
}
\caption{
Process $D_{\FDAPMone}$: modelling the FDAP with Modification 1.}
 \label{fema-mod1}
\end{figure*}
}

\newcommand{\femadiagthree}{
\begin{figure*}[!h]
\fbox{
\begin{tikzpicture}[
					thick,scale=0.6, every node/.style={transform shape},
					outline/.style={draw,thick,fill=black!3, minimum height=0.75cm,minimum width=1.5cm},
                    outline/.default=black,
                    haup/.style={thick,->},
                    hauq/.style={thick,}
                    ]
\def\myscl{2}
\node (N0) at (0,12) {};
\node (A0) at (1*\myscl,9) {};
\node [outline] (N1) at (1*\myscl,12) {\tbox{1. Disaster strikes}};
\node [outline] (N2) at (0,10) {\tbox{2. Disaster identified}};
\node [outline] (N3) at (0,8) {\tbox{3. Damage assessment}};
\node [outline] (N4) at (0,6) {\tbox{4. Gov't review damage}};
\node [outline] (N5) at (0,4) {\tbox{5. Gov'r assess response resources}};
\node [outline] (N7) at (2*\myscl,6) {\tbox{7. Region submits MDD+FA request}};
\node [outline] (N8) at (6*\myscl,6) {\tbox{8. President reviews request}};
\node [outline] (N9) at (6*\myscl,9) {\tbox{9. President declares disaster}};
\node [outline] (N10) at (6*\myscl,12) {\tbox{10. FEMA support delivered}};
\node [outline] (N6) at (2*\myscl,2) {\tbox{6. State only disaster response}};
\node [outline] (N11) at (6*\myscl,2.5) {\tbox{11. Independent Audit}};
\draw [hauq] (N1) -- (0,12) -- (N2);
\myarr{N0}{N2}{\tiny$\dprec$}
\myarr{N2}{N3}{\tiny$\dsucc$}
\myarr{N3}{N4}{\tiny$\dsucc$}
\myarr{N4}{N5}{\tiny$\dprec$}
\myarr{N7}{N8}{\tiny$\dsucc$}
\myarr{N8}{N9}{\tiny$\dweakresp$}
\myarr{N9}{N10}{\tiny$\dsucc$}
\draw [hauq] (N1) -- (1*\myscl,9) -- (N9);
\myarr{A0}{N9}{\tiny$\dprec$}
\myarr{N5}{N7}{\tiny$\dprec$}
\myarr{N5}{N6}{\tiny$\dprec$}
\node [draw,circle,minimum width=.3cm,fill=white] (Cen) at (1.3*\myscl,4) {\tiny{orresp}};
\draw [haup] (N5) -- (Cen);
\draw [haup] (Cen) -- (N7);
\draw [haup] (Cen) -- (N6);
\node at (1*\myscl,12.75) {\tiny$\dinit$};
\draw[thick] (N4) -- (2,6) -- (2,8) -- (6,8) -- (6,2.5) -- (5.5*\myscl,2.5);
\node (Z0) at (6,2.5) {};
\myarr{Z0}{N11}{\tiny$\dresp$}
\myarr{N11}{N8}{\tiny$\dprec$}
\end{tikzpicture}
}
\caption{Process $D_{\FDAPMtwo}$: modelling the FDAP with Modification 2.}
 \label{fema-mod2}
\end{figure*}
}

\newcommand{\femadiagfour}{
\begin{figure*}[!h]
\fbox{
\begin{tikzpicture}[
					thick,scale=0.6, every node/.style={transform shape},
					outline/.style={draw,thick,fill=black!3, minimum height=0.75cm,minimum width=1.5cm},
                    outline/.default=black,
                    haup/.style={thick,->},
                    hauq/.style={thick,}
                    ]
\def\myscl{2}
\node (N0) at (0,12) {};
\node (A0) at (1*\myscl,9) {};
\node (T0) at (0.9*\myscl,12) {};
\node [empty] (T1) at (0.9*\myscl,6) {};
\node [outline] (N1) at (1*\myscl,12) {\tbox{1. Disaster strikes}};
\node [outline] (N2) at (0,10) {\tbox{2. Disaster identified}};
\node [outline] (N3) at (0,8) {\tbox{3. Damage assessment}};
\node [outline] (N4) at (0,6) {\tbox{4. Gov't review damage}};
\node [outline] (N5) at (0,4) {\tbox{5. Gov'r assess response resources}};
\node [outline] (N7) at (2*\myscl,6) {\tbox{7. Region submits MDD+FA request}};
\node [outline] (N8) at (6*\myscl,6) {\tbox{8. President reviews request}};
\node [outline] (N9) at (6*\myscl,9) {\tbox{9. President declares disaster}};
\node [outline] (N10) at (6*\myscl,12) {\tbox{10. FEMA support delivered}};
\node [outline] (N6) at (2*\myscl,2) {\tbox{6. State only disaster response}};
\draw [hauq] (N1) -- (0,12) -- (N2);
\myarr{N0}{N2}{\tiny$\dprec$}
\myarr{N2}{N3}{\tiny$\dsucc$}
\myarr{N3}{N4}{\tiny$\dsucc$}
\node [empty] (T0) at (0.9*\myscl,11.65) {};
\node [hauq] (T1) at (0.9*\myscl,6) {};
\draw [hauq] (T0) -- (0.9*\myscl,6);
\draw [hauq] (0.9*\myscl,6) -- (N5);
\myarr{T1}{N5}{\tiny$\dprec$}
\myarr{N7}{N8}{\tiny$\dsucc$}
\myarr{N8}{N9}{\tiny$\dweakresp$}
\myarr{N9}{N10}{\tiny$\dsucc$}
\draw [hauq] (N1) -- (1*\myscl,9) -- (N9);
\myarr{A0}{N9}{\tiny$\dprec$}
\myarr{N5}{N7}{\tiny$\dprec$}
\myarr{N5}{N6}{\tiny$\dprec$}
\node [draw,circle,minimum width=.3cm,fill=white] (Cen) at (1.3*\myscl,4) {\tiny{orresp}};
\draw [haup] (N5) -- (Cen);
\draw [haup] (Cen) -- (N7);
\draw [haup] (Cen) -- (N6);
\node at (1*\myscl,12.75) {\tiny$\dinit$};
\end{tikzpicture}
}
\caption{Process $D_{\FDAPMthree}$: modelling the FDAP with Modification 3.}
 \label{fema-mod3}
\end{figure*}
}

\newcommand{\declarativetable}{
\begin{table}[!h]
\begin{tabular}{l|l} \hline
Constraint & Description \\ \hline
$\dinit(a)$ & An execution of the system must begin with activity $a$ \\
$\dprec(a,b)$ & If activity $b$ occurs then it must have been preceded by activity $a$ \\
$\dresp(a,b)$ & If activity $a$ occurs then activity $b$ must occur sometime later \\
$\dsucc(a,b)$ & Activity $a$ occurs if and only if activity $b$ occurs sometime later\\
$\dweakresp(a,b)$ & If activity $a$ occurs then $b$ can happen sometime later but does not have to\\
$\dorresp(a,(b,c))$ & If activity $a$ occurs then at least one of $b$, $c$ must occur sometime after it. \\ \hline
\end{tabular}
\caption{Declarative constraints used in our model}
\label{tabledeclarative}
\end{table}
}

\title[A simple declarative model of the Federal Disaster Assistance Policy]{A simple declarative model of the Federal Disaster Assistance Policy -- modelling and measuring transparency}
\author{Mark Dukes}
\address{School of Mathematics and Statistics, University College Dublin, Dublin 4, Ireland.}

\begin{document}
\begin{abstract}
In this paper we will provide a quantitative analysis of a simple model of the Federal Disaster Assistance policy from the viewpoint of three different stakeholders. This quantitative methodology is new and has applications to other areas such as business and healthcare processes. The stakeholders are interested in process transparency but each has a different opinion on precisely what constitutes transparency. We will also consider three modifications to the Federal Disaster Assistance policy and analyse, from a stakeholder viewpoint, how stakeholder satisfaction changes from process to process. This analysis is used to rank the favourability of four policies with respect to all collective stakeholder preferences.
\end{abstract}
\maketitle

\section{Introduction}\label{sectionone}
The aim of this paper is to apply techniques developed in the papers 
\cite{paper2} and \cite{paper1} to a simple model of the United States' Federal Disaster Assistance policy (FDAP). 
This policy, part of the Stafford Act, is a mechanism by which a presidential major disaster declaration 
triggers financial and physical assistance through the Federal Emergency Management Agency (FEMA)~\cite{Stafford,finedetails,critical_issues}.
In this study we will consider several types of transparency stakeholders in relation to the policy and also consider several potential modifications to this federal policy.
We will show how a comparative analysis of stakeholders in a policy-execution process can be modelled and achieved, and quantitatively examine the effect of policy modifications on stakeholder utilities.

We use the term {\it{policy-execution process}} to represent the way in which a given policy can be executed, much in the same way that a stochastic process is a process that evolves with respect to some underlying random variable. 
This is very different to the concept of Lasswell's policy cycle (or policy process)~\cite{lasswell,birkland} that exists in the policy and social sciences and concerns the formation cycle of a policy. 
Quantitative methodology in that area of the policy cycle has tended to use two types of  methods.
The first of these are text mining methods, such as {\it{term-frequency inverse-document-frequency}}, that feature in information retrieval~\cite{introIR}. 
In this instance authors typically apply this method to some chosen representative of a policy such as key phrases that appear within a policy document, or a secondary source such as debates concerning a particular policy~\cite{cross}.
The second method is social network analysis~\cite{sna}  as  applied to some chosen representative (e.g. debate keywords, key phrases) of a policy~\cite{leifeld,ba}.
The notion of policy cycle is not relevant to this paper or to our approach.

In the literature on disaster management policy research~\cite{schneider},
many studies that deal with the evaluation of such policies tend to concentrate on the affected parties~\cite{drakes,Zhang} or an analysis on the public's response to such a disaster~\cite{coles,Hu}. 
A slightly different avenue of research that is more theoretical in spirit concerns statistical models such as the Policy Modeling Research Consistency (PMC) index~\cite{pmcoriginal,china}.
These statistical models typically follow a text mining approach to evaluate the effectiveness of policies, as explained in the previous paragraph.
Disaster management papers that address the FDAP specifically include~\cite{kousky,assessingdist,rivera2019}, however none of these are directly related to the topic of this paper.

The concept of a mathematical model of a policy-execution process is new.
This was inspired, in part, by research in the area of business process management \cite{fahland}, \cite{vda1}, \cite{vda2}.
The model we use incorporates the dynamics of the policy-execution process. 
This is very different in spirit to the text mining and social network analysis approaches described above that apply statistics to representatives of policy documents.
Moreover, there is no related quantitative modelling approaches to compare it to. 
One modelling approach that bears similarities to ours is the Institutional Grammar~\cite{frantzsiddiki}, although currently it does not incorporate tools to assign and quantitatively compare utilities.
See also \cite{granger} and \cite{thompson}.
Measuring stakeholder utility for such models is a separate matter and our derivation of the utility was inspired, also in part, by the notion of entropy in information theory.
An application of this method to healthcare processes was given in \cite[Sections 4 \& 5]{paper2}.

We view a policy-execution process as a collection of events that are related to one another through a collection of {\it rules}.
A key difference between our viewpoint and a more conventional consideration of discrete event systems \cite{DES} is that the relationships between events are not necessarily causal. 
Two events can be related by rules such as `event A can happen only if event B has already occurred' or `event A can not happen before event B but does not need to'.
Attempting to model a policy by a discrete event system becomes unrealistic and a new modelling paradigm is required to encode such subtle dependencies.
This was accomplished in the first paper on this topic \cite{paper1} and further built upon in \cite{paper2}.
In essence, it used a declarative paradigm in conjunction with linear temporal logic to provide a mathematical model for policy-execution process analysis.

In Section~\ref{sec2} we give a simplified description of the Federal Disaster Assistance policy and present it in the form of an event summary.
The summary listing will be used in Section~\ref{sec3} to translate the policy into what we call a {\it{declarative process}} or {\it{declarative model}}.
We also introduce the three different stakeholders that will be considered. 
Each of these stakeholders is interested in transparency but differ in what they consider as `good' transparency. 
These stakeholders are more stakeholder types rather than actual stakeholders and motivation for their existence is provided.

In Section~\ref{sec3} we define what we mean by a declarative model and then explain how the event summary of the FDAP can be translated into a declarative model.
We also list all the different ways in which the declarative model can `run' or be executed. It turns out that for our model of the FDAP there are 46 different ways in which the process for the FDAP can be executed.
In keeping with this line of consideration, we also introduce three modifications to the FDAP and present these as declarative models. 
This will be quite straightforward to do as each is a slight alteration of the original declarative model.

In Section~\ref{sec4} we use the language we have developed in Section 3 to translate the stakeholder preferences that were summarised in Section 2 into logical propositions. 
The purpose of doing this is so that the propositions can be used to judge whether each of the many ways in which a declarative model/system may be executed is transparent for each stakeholder, or not.

Finally, in Section~\ref{sec5} we bring together all of these considerations and provide a quantitative analysis of each of the different declarative processes in the eyes of each of the three different transparency stakeholders.
This allows us to quantify and characterise the level of transparency for each of the four policies, the original FDAP and the three modifications, 
with respect to the three different stakeholders.
Our analysis allows us to decide which of the processes is optimal with respect to stakeholder preferences, and allows us to rank both processes and stakeholder satisfaction in this regard.

\section{The Federal Disaster Assistance Policy}
\label{sec2}
The Federal Disaster Assistance policy outlines the mechanism to deal with a range of disasters in the US \cite{Stafford}. 
While detailed, the essence of this policy can be captured as a series of activities that happen in a somewhat linear manner, although it 
is not necessarily true that all or any of these must occur \cite{femabrief}.
The listing of activities is

\begin{itemize}
\item
A disaster strikes.
\item
The state identifies the disaster.
\item
A damage assessment is made.
\item
Government officials review the damage and determine the extent of the disaster and its impact. 
\item
Governor decides if the state has enough resources to respond to the disaster. If not, they determine the type and amount of federal assistance they need.
\item
The state/tribe/territory submits a major disaster declaration request. 
\item
The president reviews the request and determines whether the state and local governments will need federal assistance to recover from the disaster.
\item
A disaster is declared by the president.
\item
When the major disaster declaration request is approved, FEMA begins to support the disaster response with funding, supplies, and personnel. 
\end{itemize}

Let us describe these statements as activities in a slightly more advantageous way and introduce numbers related to the different events and activities that can occur.
\ \\[1em]
\noindent\fbox{
\begin{minipage}{0.98\textwidth}
\centerline{\underline{The FDAP Event Summary}}
\begin{minipage}[t]{0.91\textwidth}
\begin{enumerate}
\item[(1)] A disaster strikes.
\item[(2)] The state identifies the disaster
\item[(3)] A damage assessment is made.
\item[(4)] Government officials review the damage and determine the extent of the disaster and its impact. 
\item[(5)] Governor decides if the state has enough resources to respond to the disaster.  If not then (7) then they submit a major disaster declaration for federal assistance request.
If yes then (6) state disaster response occurs. 
It may be the case that both (6) and (7) occur.
\item[(8)] The president reviews the request and determines whether the state and local governments will need federal assistance to recover from the disaster.
\item[(9)] A disaster is declared by the president.
\item[(10)] When the major disaster declaration request is approved, FEMA begins to support the disaster response with funding, supplies, and personnel. 
\end{enumerate}
\end{minipage}
\end{minipage}}

\medskip
The FDAP summary is a linear listing of the process but does not mention all actions, whether redundant or not, that can occur.
It might be the case that introducing a new numbered action, such as (6), is necessary in order to provide a more complete model.
The ways in which the numbered events in the above summary can occur are somewhat subtle. 
For example, a disaster may strike that wipes out all of humanity, so that the other activities never have an opportunity to occur. 

In this way activity (1) can occur by itself without other activities occurring. 
However activity (2) can only happen once activity (1) has occurred. Another way to say this is that {\em an occurrence of (2) must be preceded by an occurrence of activity (1)}.
It might be the case that every pair/triple of activities have similar interdependencies, and these dependencies could be existential or temporal in nature.

In order to faithfully model the FDAP, a detailed list of all such existential and  temporal interdependencies between collections of activities is required. 
We will return to do this in Section~\ref{sectionltl} after some necessary conventions have been introduced.
To motivate and discuss our considerations, we will now give a description of the different transparency stakeholders types.
For now the description is textual rather than symbolic, but once we have introduced the necessary technical machinery we will provide a symbolic description.

The notion of transparency in relation to a process is a subjective one. 
However, when a process is executed there are certain events/activities that a stakeholder may associate with being {\it{transparent}} in nature, and these may differ from stakeholder to stakeholder. 
Indeed it may even be the case that certain sequences of activities occurring are deemed transparent, whereas those same activities occurring in a different order are not.

\begin{description}
\item[Stakeholder 1 (Lightweight governance stakeholder)] 
This stakeholder type has a prescribed set of activities that they associate with transparency.
This stakeholder then judges an execution of the process to be good if, during the execution of the system, at least one type of review or assessment activity has occurred. 
In this instance this stakeholder is allowing for a rather tenuous view of transparency, perhaps for outward facing reasons in that any instance of one of their transparency activities is enough 
to `sell' an execution of the process that contains it.
An example of such a lightweight governance stakeholder might be the Federal Emergency Management Agency itself, or any body that wishes to show the agency acts with high regard to transparency.
\item[Stakeholder 2 (Strong governance stakeholder)] 
This stakeholder judges an execution of the process to be good if, during the execution of the system, all of the activities in the process that one could categorize as `review' or `assessment' activities occur.
This stakeholder takes a more serious view to that of Stakeholder 1 in that they would like to see all activities they associate with transparency occurring.
An example of such a stakeholder would be any collective that represents the opposition to the current government, such as a political party or media outlet, and is motivated to find issues with transparency with respect to the policy.
\item[Stakeholder 3 (Reasonable governance stakeholder)] 
This stakeholder takes a more considered view to that of the other two stakeholders in that their notion of transparency depends on an ordered sequence of activities occurring whereby there is some notion of review-before-action.
This stakeholder judges an execution of the process to be good if, during the execution of the system, either
\begin{itemize}
\item ahead of any direction at the state level, the governor or local government have reviewed or assessed information on the disaster so far, or
\item ahead of an independent audit, the governor or local government have reviewed or assessed information on the disaster so far.
\end{itemize}
The U.S. Government Accountability Office would be an example of such a stakeholder.
\end{description}

We will consider the FDAP from the point of view of these stakeholders and quantify, 
through utility measures, how each of the stakeholders perceives the FDAP and the three different modifications.

\section{A declarative model of the FDAP}
\label{sec3}
In this section we will first explain what we mean by a {\it{declarative model}} or {\it{declarative process}}.
Following that we will present our declarative model of the FDAP that we will refer to as $D_{\FDAP}$.

\subsection{Declarative models}
A declarative process is a process in which a collection of activities may occur so long as the execution of these 
activities always satisfies a listing of constraints. 
These constraints are declarations specifying when activities may/must/cannot/need-to occur with respect to one anther. 
The collection of constraints may be empty in which case all activities are free to happen as they please. 
We record the execution of a declarative process as a sequence of the activities in the order they occur and this sequence is called a {\it{trace}}.
For example, if there are four activities $\Sigma=\{A_1,A_2,A_3,A_4\}$ and activity $A_2$ occurs first, followed by $A_1$, and then $A_4$ we record this as the trace $\tau=(A_2,A_1,A_4)$.
Sometimes we will refer to a trace as a {\it valid trace} simply to emphasise that there is some notion of satisfiability in the background.

As with any new system, let us mention a few points to avoid potential confusion. 
A trace contains no `times' as which activities occur, it only lists the order in which activities occur. Not all activities have to occur in a trace.
For the system we will model each activity occurs no more than once, so every entry in a trace is unique. 
We settle on the convention that two activities cannot occur at the same time. If some argument is given for why they can, 
then this will be remedied by a consideration of events at the millisecond level.

The precise technical definition, given in \cite{paper1}, for a {\it{declarative process}} is a pairing $(\Sigma,\Const)$ where $\Sigma$ is a set of activities and $\Const$ is a set of constraints on the activities in $\Sigma$. 
The constraints in $\Const$ are similar to those described in italics in Section~\ref{sec2}.
The outstanding issue is how do we formalise modal and temporal constraints on collections of activities? 
In order to formally do this, one needs to look to an area of logic called {\em linear temporal logic} (LTL).
This is an extension to standard {\em propositional logic} that allows for the notion of time to enter the picture.
In order to make this paper accessible to a broader audience, we will avoid introducing LTL and its symbols. 
Instead, we will give high-level descriptions of the constraints that are used in our model and introduce the necessary concepts in a way 
so as to keep this paper self-contained. 

\declarativetable

We have already seen an example of an LTL constraint in Section~\ref{sec2}. This was the constraint $\dprec$.
In Table~\ref{tabledeclarative} we list the declarative constraints that feature in this paper along with their definitions.
If we have a listing $\tau$ of activities in a system and a set of constraints $\Const$, then we say $\tau$ {\it{satisfies}} $\Const$ if the listing $\tau$ satisfies each of the constraints listed in $\Const$. 
For example, the sequence $\tau=(A_2,A_1,A_4)$ satisfies $\Const = \{ \dprec(A_2,A_4), \dresp(A_3,A_2)
\}$. 
However $\tau=(A_2,A_1,A_4)$ does {\it{not}} satisfy $\Const = \{ \dresp(A_1,A_2), \dsucc(A_2,A_4)\}$ since $A_1$ appears in $\tau$ but it $A_2$ does not occur after $A_1$.

\subsection{The \FDAP}
\label{sectionltl}
Consider the FDAP as described in Section~\ref{sec2} in conjunction with the constraints given in Table~\ref{tabledeclarative}.
We can describe how the events are related to one another as the following set of declarative constraints:

$$\Const_{FDAP} = \left\{ \begin{array}[c]{l} 
								\dprec(1,2), \dprec(1,9), \dsucc(2,3), \dsucc(3,4), \dprec(4,5), \dprec(5,6),
								 \dprec(5,7),\\ 
								\dorresp(5,(6,7)), \dsucc(7,8), \dweakresp(8,9), \dsucc(9,10)  \end{array}
\right\}.$$
\femadiagone
To explain how these constraints were arrived at, look at the FDAP event summary. We see that activity (2) {\it The state identifies a disaster} can only occur if (1) {\it A disaster strikes}.
Activity (2) cannot happen without activity (1) having occurred. 
We saw there that it is possible for activity (1) to occur and for activity (2) to not occur.
The declarative constraint that summarises this relationship is $\dprec(1,2)$, that activity (1) must precede activity (2). All other precedence constraints in $\Const_{FDAP}$ can be reasoned in this same way.

Consider next the relationship between activity (2) {\it The state identifies a disaster} and activity (3) {\it A damage assessment is made}. If activity (2) occurs then activity (3) must occur as a result, which translates into the constraint $\dresp(2,3)$. Similarly, activity (3) can occur only if activity (2) has occurred, and this translates into the constraint $\dprec(2,3)$. The combination of these two constraints $\dresp(2,3) \wedge \dprec(2,3)$ is $\dsucc(2,3)$. Here the symbol $\wedge$ represents {\it{logical and}} in propositional logic. All other instances of $\dsucc$ in the constraint set  $\Const_{FDAP}$ can be reasoned in this way.

The two remaining constraints that feature in the list $\Const_{FDAP}$ that have yet to be justified are $\dorresp(5,(6,7))$ and $\dweakresp(8,9)$.
The constraint $\dorresp(5,(6,7))$, when one refers to the definition of $\dorresp$ in Table~\ref{tabledeclarative}, is essentially a direct translation of what is mentioned in the FDAP event summary in Section 2.
The constraint $\dweakresp(8,9)$ might seem somewhat out of place at first sight, 
but the reason for the weak response (as opposed to response) is that on the president reviewing the request, it could be the case that he does not declare a disaster. 
If activity (8) happens then activity (9) can occur, but does not have to.

We refer to this as the declarative process $D_{\FDAP} = (\{1,2,\ldots, 10\}, \Const_{FDAP})$ where the set $\{1,2,\ldots,10\}$ is the list of all the possible activities in the process.
Figure~\ref{fema-as-is} illustrates these constraints in the form of an edge-labelled graph where the inter-dependencies amongst actions in relation to constraints is easier to check.
There are a finite number of ways in which this declarative process can be executed. 
Each of these corresponds to a (valid) trace, and it is these valid traces and subsets thereof that play a central role in the calculation of stakeholder utility measures.
A listing of all the 46 valid traces for this declarative process is given in Table~\ref{validtracestable}. 
Determining this precise number is far from trivial and it was obtained using software specifically designed for declarative processes (this is discussed in \cite{paper1} and the software for this paper can be found in \cite{dukessoftware}).
An interesting remark is that if we were to consider a declarative process on 10 activities that has no constraints (i.e. the constraint set is empty), then the number of valid traces 
of such a system is 9,864,102 (see \cite{paper2}).

\begin{table}[!h]
\scriptsize
$$\begin{array}{|c|c|c|c|} 
 \hline
\begin{array}[t]{ll}
1.&  \epsilon \\
2.& (1) \\
3.& (1, 9, 10) \\ 
4.& (1, 2, 3, 4) \\
5.& (1, 2, 3, 4, 5, 6) \\
6.& (1, 2, 3, 4, 9, 10) \\
7.& (1, 2, 3, 9, 4, 10) \\
8.& (1, 2, 3, 9, 10, 4) \\
9.& (1, 2, 9, 3, 4, 10) \\
10.& (1, 2, 9, 3, 10, 4) \\
11.& (1, 2, 9, 10, 3, 4) \\
12.& (1, 9, 2, 3, 4, 10) \\
\end{array} & \begin{array}[t]{ll}
13.& (1, 9, 2, 3, 10, 4) \\
14.& (1, 9, 2, 10, 3, 4)\\
15.& (1, 9, 10, 2, 3, 4) \\
16.& (1, 2, 3, 4, 5, 7, 8) \\
17.& (1, 2, 3, 4, 5, 6, 7, 8) \\
18.& (1, 2, 3, 4, 5, 7, 6, 8) \\
19.& (1, 2, 3, 4, 5, 7, 8, 6) \\
20.& (1, 2, 3, 4, 5, 6, 9, 10) \\
21.& (1, 2, 3, 4, 5, 9, 6, 10) \\
22.& (1, 2, 3, 4, 5, 9, 10, 6) \\
23.& (1, 2, 3, 4, 9, 5, 6, 10)\\
24.& (1, 2, 3, 4, 9, 5, 10, 6) \\
\end{array} & \begin{array}[t]{ll}
25.& (1, 2, 3, 4, 9, 10, 5, 6) \\
26.& (1, 2, 3, 9, 4, 5, 6, 10) \\
27.& (1, 2, 3, 9, 4, 5, 10, 6) \\
28.& (1, 2, 3, 9, 4, 10, 5, 6) \\
29.& (1, 2, 3, 9, 10, 4, 5, 6) \\
30.& (1, 2, 9, 3, 4, 5, 6, 10) \\
31.& (1, 2, 9, 3, 4, 5, 10, 6) \\
32.& (1, 2, 9, 3, 4, 10, 5, 6) \\
33.& (1, 2, 9, 3, 10, 4, 5, 6) \\
34.& (1, 2, 9, 10, 3, 4, 5, 6) \\
35.& (1, 9, 2, 3, 4, 5, 6, 10) \\
36.& (1, 9, 2, 3, 4, 5, 10, 6) \\
\end{array} & \begin{array}[t]{ll}
37.& (1, 9, 2, 3, 4, 10, 5, 6) \\
38.& (1, 9, 2, 3, 10, 4, 5, 6) \\
39.& (1, 9, 2, 10, 3, 4, 5, 6) \\
40.& (1, 9, 10, 2, 3, 4, 5, 6) \\
41.& (1, 2, 3, 4, 5, 7, 8, 9, 10) \\
42.& (1, 2, 3, 4, 5, 6, 7, 8, 9, 10) \\
43.& (1, 2, 3, 4, 5, 7, 6, 8, 9, 10) \\
44.& (1, 2, 3, 4, 5, 7, 8, 6, 9, 10) \\
45.& (1, 2, 3, 4, 5, 7, 8, 9, 6, 10) \\
46.& (1, 2, 3, 4, 5, 7, 8, 9, 10, 6) 
\end{array} \\ \hline
\end{array} $$
\caption{Valid traces for the declarative process $D_{\FDAP}$}
\label{validtracestable}
\end{table}

Let us now consider three different modifications to the FDAP and the declarative models which correspond to these modifications.
From a policy perspective, these modifications are easy to justify. 
The first modification considers the variant of the FDAP in which no presidential discretion is allowed in declaring a disaster.
Perhaps it has been the case that a president abused his power for personal gain though declaring a disaster early.
This modification is used to consider how stakeholder transparency compares with the original FDAP if such a presidential declaration is forbidden without reason.
In order to enact this modification, no change to the action set is required, only to the set of constraints.

The second modification to the FDAP seeks to model what happens if a new (independent, non-governmental) auditing activity is introduced into the process.
This new process must happen before a president reviews a request for assistance, and is triggered as a result of a government review into any damage. 
The third modification seeks to understand how the dynamics of the policy-execution process changes if less restrictions are placed on, and more power is given to, the state governor.
Indeed, any arbitrary modification could also be easily considered and accommodated. 
However we have attempted to choose modifications that are realistic and related to the stakeholders we consider.

\subsection*{Modification 1} {\em No presidential discretion.}
In this modification of the process $\FDAP$, the president is not allowed to simply declare a disaster (as he has done on occasion). 
Instead he must wait for the formal submission of a request for federal assistance. 
This can be modelled declaratively by replacing the constraint $\dprec(1,9)$ in $D_{\FDAP}$ with $\dprec(8,9)$. More formally, 
the new declarative process is
$$D_{\FDAPMone} = (\{1,2,\ldots,10\}, \Const_{\FDAP} - \{\dprec(1,9)\} \cup \{ \dprec(8,9)\}),$$
where the $-$ and $\cup$ are the mathematical operators to change the constraint set in the necessary way.
Figure~\ref{fema-mod1} contains an illustration of these constraints.
\femadiagtwo
The valid traces for $D_{\FDAPMone}$ are 14 in number and are illustrated in Table~\ref{tablemodified}.

\subsection*{Modification 2}
{\em Independent audit required.}
An independent audit of the damage impact and resources required is to be conducted ahead of any presidential consideration.
The purpose of this is to improve transparency. 
This can be modelled by the constraints in $\Const_{\FDAP}$ together with 
$\dresp(4,11)$ and $\dprec(11,8)$ where action 11 is
\ \\[1em]
\noindent\fbox{\fbox{
\begin{minipage}{0.96\textwidth}
\begin{minipage}[t]{0.91\textwidth}
\begin{enumerate}
\item[(11)] Independent audit of damage impact and resources required is contracted.
\end{enumerate}
\end{minipage}
\end{minipage}
}}
\medskip

We formally define this as the declarative process 
$$D_{\FDAPMtwo} = (\{1,2,\ldots,10,11\}, \Const_{\FDAP} \cup \{ \dresp(4,11),\dprec(11,8)\}).$$
\femadiagthree
See Figure~\ref{fema-mod2} for an illustration of these constraints. There are 144 valid traces for this modification and these can be found in \cite{dukessoftware}.

\subsection*{Modification 3}
{\em State governor may act unilaterally.}
In this modification the state governor does not have to wait on either the damage assessment or the governmental review of the damage. 
This can be modelled by removing the constraint $\dprec(4,5)$ in $ \Const_{\FDAP}$ and introducing the constraint $\dprec(1,5)$:
$$D_{\FDAPMthree} = (\{1,2,\ldots,10\}, \Const_{\FDAP} - \{ \dprec(4,5)\}\cup \{ \dprec(1,5)\}).$$
\femadiagfour
See Figure~\ref{fema-mod3} for an illustration of these constraints.
The are 852 valid traces for this modification and these can be found in \cite{dukessoftware}.

\begin{table}[!b]
$$\begin{array}{|c|c|} \hline
\begin{array}[t]{ll}
1. & \epsilon \\
2. & (1) \\
3. & (1, 2, 3, 4) \\
4. & (1, 2, 3, 4, 5, 6) \\
5. &  (1, 2, 3, 4, 5, 7, 8) \\
6. &  (1, 2, 3, 4, 5, 6, 7, 8) \\
7. & (1, 2, 3, 4, 5, 7, 6, 8)
\end{array} & \begin{array}[t]{ll}
8. & (1, 2, 3, 4, 5, 7, 8, 6) \\
9. & (1, 2, 3, 4, 5, 7, 8, 9, 10) \\
10. & (1, 2, 3, 4, 5, 6, 7, 8, 9, 10) \\
11. &  (1, 2, 3, 4, 5, 7, 6, 8, 9, 10) \\
12. &  (1, 2, 3, 4, 5, 7, 8, 6, 9, 10) \\
13. & (1, 2, 3, 4, 5, 7, 8, 9, 6, 10) \\
14. & (1, 2, 3, 4, 5, 7, 8, 9, 10, 6)
\end{array} \\ \hline
\end{array} $$
\caption{Valid traces for the declarative process $D_{\FDAPMone}$}
\label{tablemodified}
\end{table}

\section{Modelling stakeholder preferences}
\label{sec4}
A stakeholder, in the context of a policy-execution process, is someone who has a clearly defined opinion about the execution of such a system.
To clear up any potential disambiguation, stakeholders play no part in the execution of a policy-execution process or system, and do not impact it in any way.
They are essentially part of the audience to the different possible executions of events.

At the simplest level, and this is the one which we will consider, a stakeholder $S$ will hold a binary view of whether the execution of a system is favourable. 
This assignment was considered in the papers \cite{paper2} and \cite{paper1}.
To every trace $\tau$ we associate a value $S(\tau)$ in $\{0,1\}$ signifying whether a stakeholder judges $\tau$ to be not-favourable/favourable.

We next describe the preferences of the three stakeholders introduced in Section~\ref{sec2} using LTL expressions.
More precisely, we will translate a stakeholder's preference into an LTL expression, against which every trace of the declarative model/process is checked. 
The outcome of this will be a number $S(D)$ that signifies the sum of the binary values for stakeholder $S$ over all valid traces $\tau$ of the system $D$. 
The precise values of $S(D)$ for the different declarative processes and stakeholders will be presented in the subsequent section.

\subsection*{Stakeholder 1}
This stakeholder judges an execution of the process to be good if, during the execution of the system, there is at least some type of review or assessment activity that has occurred. 
To model this we must pay attention to the activities in the process under consideration. In the processes we consider these will either be $\{1,\ldots,10\}$ or $\{1,\ldots,10,11\}$. 
If the underlying activity set is $\{1,\ldots,10\}$, then we will consider a trace $\tau$ to be good for this stakeholder if at least one of the review/assessment activities 4, 5, or 8 make an appearance:
$$(4 \in \tau) \mbox{ or } (5 \in \tau) \mbox{ or }(8 \in \tau).$$
Here the logical symbol $\in$ is shorthand for `is in'.
If the underlying activity set is $\{1,\ldots,10,11\}$, the we will consider a trace $\tau$ to be good for this stakeholder if at least one of the review/assessment activities 4, 5, 8, or 11 make an appearance:
$$(4 \in \tau) \mbox{ or } (5 \in \tau) \mbox{ or }(8 \in \tau)\mbox{ or }(11 \in \tau).$$
Both of these conditions can be consolidated into the single statement that a trace $\tau$ is good for this stakeholder if
\begin{align}\label{stakecondone}
\tau \mbox{ satisfies the LTL constraint } (4 \in \tau) \vee (5 \in \tau) \vee (8 \in \tau) \vee (11 \in \tau),
\end{align}
where $\vee$ is the symbol for {\it{logical or}}.

\subsection*{Stakeholder 2}
This stakeholder judges an execution of the process to be good if, during the execution of the system, all of the activities in the process that one could categorize as `review' or `assessment' activities occur. Just as we did with Stakeholder 1, we have two cases depending on the underlying set of activities:
\begin{enumerate}
\item[(A)] If the underlying set of activities of the process is $\{1,\ldots,10\}$, then a trace $\tau$ is good for this stakeholder if
\begin{align}\label{stakecondtwoA}
\tau \mbox{ satisfies the LTL constraint } (4 \in \tau) \wedge (5 \in \tau) \wedge (8 \in \tau). 
\end{align}
\item[(B)] If the underlying set of activities of the process is $\{1,\ldots,11\}$, then a trace $\tau$ is good for this stakeholder if
\begin{align}\label{stakecondtwoB}
\tau \mbox{ satisfies the LTL constraint } (4 \in \tau) \wedge (5 \in \tau) \wedge (8 \in \tau) \wedge (11 \in \tau). 
\end{align}
\end{enumerate}

\subsection*{Stakeholder 3}
This stakeholder judges an execution of the process to be good if, during the execution of the system, either
\begin{itemize}
\item ahead of any direction at the state level, the governor or local government have reviewed or assessed information on the disaster so far, or
\item ahead of an independent audit, the governor or local government have reviewed or assessed information on the disaster so far.
\end{itemize}
For the first of these, we have that when activities 7 or 6 occur, then they must be preceded by both 4 and 5. 
For the second, if activity 11 occurs then it too must be preceded by both 4 and 5.
\begin{enumerate}
\item[(A)] If the underlying set of activities of the process is $\{1,\ldots,10\}$, then a trace $\tau$ is good for this stakeholder if
\begin{align}\label{stakecondthreeA}
\tau \mbox{ satisfies the LTL constraint } 
\begin{array}[t]{l}
(\dmustexist(6) \wedge ( \dprec(4,6) \vee \dprec(5,6))) \\
\vee
(\dmustexist(7) \wedge ( \dprec(4,7) \vee \dprec(5,7))).
\end{array}
\end{align}
\item[(B)] If the underlying set of activities of the process is $\{1,\ldots,11\}$, then a trace $\tau$ is good for this stakeholder if
\begin{align}\label{stakecondthreeB}
\tau \mbox{ satisfies the LTL constraint } 
\begin{array}[t]{l} (\dmustexist(6) \wedge ( \dprec(4,6) \vee \dprec(5,6)))  \\
\vee
(\dmustexist(7) \wedge ( \dprec(4,7) \vee \dprec(5,7))) \\
\vee
(\dmustexist(11) \wedge ( \dprec(4,11) \vee \dprec(5,11))).
\end{array}
\end{align}
\end{enumerate}

While the semantics of these LTL expressions might appear daunting to some, the reader will be relieved to know that 
any further understanding of them is not required in order to grasp the remainder of this paper. 
They are simply the expressions used in checking, at a coding level, whether traces are favourable or not favourable to stakeholders.
Our main purpose in presenting these LTL expressions is to show how they were derived.

\section{Results}
\label{sec5}
Now that we have have translated our four processes into declarative systems, and determined LTL expressions for the stakeholders in these systems, we are in a position to calculate stakeholder utilities.
This paper would be incomplete without a discussion of the metrics used for the stakeholder utilities and the theory upon which they are based.
However, we do not wish to distract from the topic at hand and force the reader into a technical alleyway. 
The following explanation attempts to balance these two points and the interested reader is encouraged to consult \cite{paper1} and \cite{paper2} for further and more in-depth details.

Consider a declarative process $D$ and stakeholder $S$. 
Suppose there are $\valid(D)$ valid traces of $D$ and that $S(D)$ is the number of valid traces that the stakeholder decides are favourable.
We wish to associate a utility $u_S(D)$ that represents stakeholder $S$'s satisfaction with $D$. 
Should every valid trace of $D$ be favourable for the stakeholder, then we wish $u_S(D)$ to be 1.
If no valid traces of $D$ are favourable for the stakeholder, then we wish $u_S(D)$ to be 0.
The utility $u_S(D)$ takes values in-between 0 and 1 (inclusive) and signifies the stakeholders satisfaction with $D$. 

However, essential scaling arguments steer this value away from being the simple proportion $S(D)/\valid(D)$. 
The subtlety is for the utility increase when the number of favourable traces increases by one to become less and less as the number of favourable traces increases.
This means the utility increase when we move from, say, 5 to 6 favourable traces is strictly less than the utility increase when moving from 4 to 5 favourable traces.
By extension, this means that if the number of valid traces of a system were to double, then in order to achieve the same utility the number of favourable traces would still be of a reasonable size but not quite double the number of original favourable traces.
In the papers \cite{paper1} and \cite{paper2} a mathematical derivation of the utility satisfying the properties that we deemed reasonable was given.
The derived stakeholder utility measure is 
\begin{align}
\label{equationsix}
u_S(D) & = \dfrac{\log  (1+S(D))}{\log  (1+\valid(D))}.
\end{align}

Now that we have formal descriptions for the declarative processes, and are able to classify when traces of the declarative models are favourable in the eyes of each of the stakeholders, 
we can give expressions for the stakeholders utilities for the model and its three different modifications using equation (\ref{equationsix}). 
These are summarised in Table~\ref{summarytable}.
The numbers in this table were calculated by using software designed to check declarative constraints \cite{dukessoftware}.
For each declarative process, the software first exhaustively calculates all valid traces. 
Then each of these valid traces is tested for satisfiability by the LTL expressions that represent each of the three stakeholder preferences.

\begin{table}[!h]
\begin{tabular}{|c||c||c|c|c|} \hline
Process $D$ &  $\valid(D)$  & $S_1(D)$ & $u_{S_1}(D)$ & rank \\ \hline
\FDAP &46 & 43 & 0.982869 & 3rd \\
\FDAPMone& 14 & 12 & 0.947157& 4th \\
\FDAPMtwo &144 & 141 & 0.995799& 2nd \\
\FDAPMthree &852 & 849 & 0.999478 & 1st \\ \hline \hline
Process $D$ &  $\valid(D)$  & $S_2(D)$ & $u_{S_2}(D)$ & rank \\ \hline
\FDAP &46 & 10 & 0.622806 & 4th \\
\FDAPMone& 14 & 10 & 0.885469 & 2nd \\
\FDAPMtwo &144 & 34 & 0.714394 & 3rd \\
\FDAPMthree &852 & 601 & 0.948361 & 1st \\ \hline\hline
Process $D$ &  $\valid(D)$  & $S_3(D)$ & $u_{S_3}(D)$ & rank \\ \hline
\FDAP &46 & 32 & 0.908149 & 4th \\
\FDAPMone& 14 & 11 & 0.917600 & 3rd \\
\FDAPMtwo &144 & 137 & 0.990058 & 2nd \\
\FDAPMthree &852 & 838 & 0.997548 & 1st \\ \hline
\end{tabular}
\ \\
\caption{Summary of the trace and stakeholder data for the four processes, along with the utilities for each stakeholder calculated using equation (\ref{equationsix}).\label{summarytable}}
\end{table}

This data now enables us to apply the method from \cite{paper2} and deduce the following observations.

\subsection*{Collective stakeholder-preferred process}
Section 5 in paper \cite{paper2} illustrated a method that allowed us to rank the declarative processes in decreasing order of collective stakeholder preferences, 
thereby allowing for a comparative analysis with respect to stakeholders.
To do this here, let us consider the collective stakeholder utility distances from the optimal utility (which is 1 in each case) for each of the four processes.
\begin{align*}
H(D_{\FDAP}) &= \sqrt{(1-u_{S_1}(D_{\FDAP}))^2 +  (1-u_{S_2}(D_{\FDAP}))^2 + (1-u_{S_3}(D_{\FDAP}))^2}\\
&= 0.388594.
\end{align*}
Had this $H(D_{\FDAP})$ value been 0 them the process could not be more favourable for all of the stakeholders. However, had $H(D_{\FDAP})$ taken the value $\sqrt{3}\approx 1.732$ (the square root of the number of stakeholders), then it is as unfavourable for all of the stakeholders, and could not be more so.
Doing the same for the other processes and tabulating gives the number in Table~\ref{tablex}.
\begin{table}
\begin{center}
\begin{tabular}[!h]{|c||c|c|} \hline
Process $D$ & $H(D)$ & Rank \\ \hline\hline
$D_{\FDAP}$ & 0.388594 & 4th \\
$D_{\FDAPMone}$ & 0.150664 & 2nd \\
$D_{\FDAPMtwo}$ & 0.285810 & 3rd \\
$D_{\FDAPMthree}$ & 0.051700 & 1st \\ \hline
\end{tabular}
\end{center}
\caption{\label{tablex}}
\end{table}
This reveals the listing of processes in decreasing order of collective favourability:
\begin{center}
\FDAPMthree, \FDAPMone, \FDAPMtwo, and lastly \FDAP.
\end{center}
That \FDAPMthree\ is collectively the most favourable is in this case apparent without performing the $H$-value calculations since it is the one for which the largest utility is attained for each of the users, amongst all four processes.

A policy level explanation for why this is the case is it allows a more liberal approach to the governor assessing the response resources (activity 5) while not preventing the other good-practice-from-a-transparency viewpoint activities such as activity 4. 

\subsection*{A more detailed comparison for all different cohorts of stakeholders}
Suppose that instead of considering the global optimization problem for all stakeholders, we do this for all possible subsets of stakeholders.
Are there situations in which, when we consider only one or two of the stakeholders rather than all three, some process other than $\FDAPMthree$ is optimal?
Can this analysis inform us of those collections of stakeholders that have a common optimal process and, moreover, if so then what can it reveal about the cohorts that have a prescribed process as its optimal solution?

Table~\ref{finaltable} records the reduced stakeholder utility vectors for all possible subsets of stakeholders.
The entries in the bottom row are those calculated in Table~\ref{tablex}. 
The other entries are calculated in a similar way but include only the utility values for those stakeholders that feature. For example, 
The first entry in the row for $X=\{S_1,S_3\}$ is calculated by evaluating:
$$\sqrt{(1-u_{S_1}(D_{\FDAP}))^2 + (1-u_{S_3}(D_{\FDAP}))^2}.$$
For each such subset we record in the rightmost column the optimal choice of process. 
In every case it is the third modification $\FDAPMthree$, so for this case and these stakeholders, $\FDAPMthree$ appears globally optimal.
A more in-depth analysis is not required for the processes under consideration.

\begin{table}[!h]
\begin{center}
\begin{tabular}{c|c|c} \hline
Subset & \multirow{3}{*}{$\left(H^{(X)}(D),~ H^{(X)}(D_{1}),~ H^{(X)}(D_{2}), ~ H^{(X)}(D_{3})\right)$} & Process \\
$X=\{S_{i_1},\ldots,S_{i_k}\}$ & & $J\in \{D, D_1, D_2, D_3\}$ \\
of stakeholders && that minimizes $H(J)$\\ \hline \hline
$\{S_1\}$ & (0.017131, 0.052843, 0.004201, 0.000522) & $D_3$ \\
$\{S_2\}$ & (0.377194, 0.114531, 0.285606, 0.051639) & $D_3$ \\
$\{S_3\}$ & (0.091851, 0.082400, 0.009942, 0.002452) & $D_3$ \\
$\{S_1,S_2\}$ & (0.377583, 0.126134, 0.285637, 0.051642) & $D_3$ \\
$\{S_1,S_3\}$ & (0.093435, 0.097888, 0.010793, 0.002507) & $D_3$ \\
$\{S_2,S_3\}$ & (0.388216, 0.141093, 0.285779, 0.051697)& $D_3$  \\
$\{S_1,S_2,S_3\}$ & (0.388594, 0.150664, 0.285810, 0.051700)&  $D_3$\\ \hline
\end{tabular}
\end{center}
\caption{In this table we use $D$ to refer to $D_{\FDAP}$ and $D_i$ to refer to $D_{\FDAP Mj}$ for $j=1,2,3$.
The process in the right column is the index that minimizes the sequence of values in the middle column.
\label{finaltable}}
\end{table}

\medskip

\subsection*{Naive transparency criteria makes for a satisfied stakeholder}
For each of the four processes, the utility of $S_1$ for that process is greater than the utility of each of others for that same process. 
This is explainable and to be expected since a crude and naive transparency condition defines Stakeholder 1. 
Any hint of transparency represents a contribution to the number of favourable traces for Stakeholder 1, whereas for the other stakeholders more selective criteria are used. 
It is reassuring to see this assessment jumping out from the data.

\subsection*{Utility proximity and benchmarking} 
Almost half of the utilities in Table~\ref{summarytable} are very close in value, with five of the twelve utilities having  a value between 0.98 and 1. 
This brings forth the topic of utility benchmarking -- what can be inferred about stakeholder satisfaction by announcing the utility is in a certain range or having a certain value?
The utility mentioned here is the exponent in a power-law type behavior relating the number of favourable traces to the number of valid traces, i.e. the value $u$ that satisfies $S(D) \approx \valid(D)^u$.
Thus if we are talking about those stakeholders or processes whose utility is in a certain range of values $(\alpha,\beta)$, then this translates into an inequality that gives upper and lower bounds on the number of favourable outcomes:
$$\valid(D)^{\alpha} \leq S(D) \leq \valid(D)^{\beta}.$$
To focus now on particular values, let us consider 
$$u_{S_1}(D_{\FDAPMtwo}) = 0.995799 \quad \mbox{ and }\quad  u_{S_3}(D_{\FDAPMthree}) = 0.997548.$$ 
The number of valid traces in $D_{\FDAPMthree}$ is almost 6 times as many as those in $D_{\FDAPMtwo}$ but the utilities are almost identical in value since they differ by less than $0.002$. Given this, all one can infer is that the stakeholder satisfaction is essentially the same and one cannot distinguish between these even though one is marginally bigger than the other.

However, if there is a significant difference between values then one can definitively distinguish between processes and stakeholders in a comparative way. Consider $u_{S_2}(D_{\FDAPMtwo})=0.714394$. Due to the underlying power-law behavior, it is an order of magnitude lower than $u_{S_3}(D_{\FDAPMthree}) = 0.997548$.

For example, if the number of valid traces in both $D_{\FDAPMtwo}$ and $D_{\FDAPMthree}$ were 1 million, then a utility of $0.714394$ would correspond to 19,335 favourable traces (less than one fiftieth), whereas a utility of 0.997548 corresponds to 966,691 favourable traces (almost all of them).

One other point that deserves a mention is how this method is robust in a comparative sense. 
It might be the case that there are certain traces, of length 1 say, that unintentionally satisfy a transparency constraint. For example this could be the case with trivial satisfiability of the empty trace.
Why then would this trace be marked as transparent? Should the stakeholder LTL condition not be altered to ensure they are not transparent?
The answer to this is yes, one could do that and recalculate. 
However, if a trace of small length satisfies an LTL condition, then it will most probably also satisfy the LTL constraints of the other stakeholders for the same (unintended) reason.
Because of this, the comparative utilities for each should be unaffected, even though there might be a small change to the utilities themselves.

\subsection*{Advantages and limitations of the declarative modelling approach}
An advantage to the declarative modelling approach presented in this paper is its universal simplicity. 
There is nothing disaster-specific within the analysis and so the same method could be used equally well for analysing healthcare or business processes.
Another advantage is that aspects to it could be automated rather easily, such as the decomposition of a policy into its constituent activities and the constraints relating these to one another.
The utility calculations are straightforward to automate once there are LTL expressions for the stakeholder preferences. 

However, the specification of stakeholder preferences themselves is not an easy one to automate, and 
is due to the subjective nature of how the events and activities can be viewed as {\it{good}} or {\it{bad}} in the eyes of a stakeholder.
This is one clear limitation in that it seems human input is essential in describing stakeholder preferences.

\section{Discussion}
The analysis of the stakeholders and different versions of the FDAP culminates in a rather technical way with tables of numbers.
For this reason it might seem that the original policy related questions have been cast aside, so let us address that issue here in a non-technical way.

The starting point for this paper was the following question: 
if one models the FDAP as a process of activities dependent upon one another, and considers different types of observers, then what is their degree of happiness with all of the different ways in which the FDAP can happen?
The utility analysis in Section~\ref{sec5} 
allowed us to calculate quantities that signify observer happiness with the FDAP process, thereby giving us a ranking of the observers in relation to the policy.
These values appear in Table~\ref{summarytable} in the fourth column (those rows on which FDAP is mentioned) and allow us to see that the first observer is happiest with the FDAP, followed by the third observer, followed by the second observer.

Following this, we introduced three variants to the FDAP, and asked the same question regarding observer happiness and were able to clearly rank observer happiness for each of the three variants. 
This allowed us to give an answer to the following question: 
if we have models for four different policies (the FDAP and the three variants), and consider just one of the observers, then can we rank the policies in order of happiness for this observer? 
The answer to this is yes, and the details are given by inspecting the fourth column in Table~\ref{summarytable}.

We have been able to focus on a policy and rank observers in terms of their happiness, and then focus on an observer and rank policies in terms of a given observers happiness.
Our next motivating question was: given all of these processes and observers, is it possible to talk sensibly about collective observer happiness in relation to policies and rank policies in relation 
to this collective happiness?
This was the topic of the latter part of Section~\ref{sec5} and allowed several insights into preferred policies for different collections of observers. 

\section{Conclusion and Outlook}
In this paper we have illustrated how to perform a simple declarative analysis of the Federal Disaster Assistance policy using the methods designed in \cite{paper2} and \cite{paper1}.
This allowed us to quantitatively compare three different modifications to this policy with respect to three different transparency stakeholders. 
This technique represents a new quantitative tool that can assist policy analysts in the 
redesign and refinement of policies where appeasing the views of multiple stakeholders is 
a fundamental objective.
The
stakeholder utilities in this paper are built from simple binary quantities for possible policy-executions in which a 0 represents dissatisfaction and 1 represents satisfaction. 
Although crude, these have been enough to provide for a balanced comparative analysis that seems to correspond to what would be expected for stakeholder comparisons.
There are several other more general methods by which this could be done and these will be the subject of future research in this area.

This paper is a first quantitative study of a model of the FDAP.  
It is not the intention of this article to be the definitive quantitative study.
Rather, we hope that the approach, assumptions, and results from this paper will stimulate new discussions 
about quantitative aspects of this and other policy-execution processes, while allowing other researchers to inspect, modify, and improve upon it as they please.

\end{document}